%% file: sample-sigconf.tex
\documentclass[sigconf]{acmart}
\usepackage{booktabs} % For formal tables
\usepackage{subfigure}

\setcopyright{rightsretained}
\copyrightyear{2018}
\acmYear{2018}
\setcopyright{acmcopyright}
\acmConference[EE-USAD'18]{Understanding Subjective Attributes of Data, with the Focus on Evoked Emotions 2018 Workshop}{October 22, 2018}{Seoul, Republic of Korea}
\acmBooktitle{Understanding Subjective Attributes of Data, with the Focus on Evoked Emotions 2018 Workshop (EE-USAD'18), October 22, 2018, Seoul, Republic of Korea}
\acmPrice{15.00}
\acmDOI{10.1145/3267799.3267802}
\acmISBN{978-1-4503-5978-8/18/10}
\editor{Jennifer B. Sartor}
\editor{Theo D'Hondt}
\editor{Wolfgang De Meuter}

\begin{document}
\title{What Makes Natural Scene Memorable?}
% \titlenote{Produces the permission block, and
% \subtitle{Extended Abstract}
% \subtitlenote{The full version of the author's guide is available as
%   \texttt{acmart.pdf} document}

\author{Jiaxin Lu}
%\authornote{Dr.~Trovato insisted his name be first.}
\orcid{0000-0002-8034-8247}
\affiliation{%
  \institution{Beihang University}
  \streetaddress{No.37 Xueyuan Road}
  \city{Beijing}
  \state{China}
  \postcode{100191}
}
\email{lu-jia-xin@163.com}

\author{Mai Xu}
\authornote{Corresponding author}
\affiliation{%
  \institution{Beihang University}
  \streetaddress{No.37 Xueyuan Road}
  \city{Beijing}
  \state{China}
  \postcode{100191}
}
\email{maixu@buaa.edu.cn}

\author{Ren Yang}
\orcid{0000-0003-4124-4186}
\affiliation{%
  \institution{Beihang University}
  \streetaddress{No.37 Xueyuan Road}
  \city{Beijing}
  \state{China}
  \postcode{100191}
}
\email{yangren@buaa.edu.cn}

\author{Zulin Wang}
\affiliation{%
  \institution{Beihang University}
  \streetaddress{No.37 Xueyuan Road}
  \city{Beijing}
  \state{China}
  \postcode{100191}
}
\email{wzulin@buaa.edu.cn}

\renewcommand{\shortauthors}{J. Lu et al.}

\begin{abstract}
Recent studies on image memorability have shed light on the visual features that make generic images, object images or face photographs memorable. However, a clear understanding and reliable estimation of natural scene memorability remain elusive. In this paper, we provide an attempt to answer: ``what exactly makes natural scene memorable''. Specifically, we first build LNSIM, a large-scale natural scene image memorability database (containing 2,632 images and memorability annotations). Then, we mine our database to investigate how low-, middle- and high-level handcrafted features affect the memorability of natural scene. In particular, we find that high-level feature of scene category is rather correlated with natural scene memorability. Thus, we propose a deep neural network based natural scene memorability (DeepNSM) predictor, which takes advantage of scene category. Finally, the experimental results validate the effectiveness of DeepNSM.
\end{abstract}

% \ccsdesc[300]{Human-centered computing~Human computer interaction (HCI)}


\maketitle

\input{samplebody-conf1}

\bibliographystyle{ACM-Reference-Format}
\bibliography{references}

\end{document}

%% file: samplebody-conf1.tex
\textbf{ACM Reference Format:}\\
Jiaxin Lu, Mai Xu, Ren Yang, and Zulin Wang. 2018. What Makes Natural Scene Memorable?. In Understanding Subjective Attributes of Data, with the Focus on Evoked Emotions 2018 Workshop (EE-USAD’18), October 22, 2018, Seoul, Republic of Korea. ACM, New York, NY, USA, 7 pages. https://doi.org/10.1145/3267799.3267802
\section{Introduction}
One hallmark of human cognition is the splendid capacity of recalling thousands of different images, some in details, after only a single view. In fact, not all images are remembered equally in human brain. Some images stick in our minds, while others fade away in a short time. This kind of capacity is likely to be influenced by individual experiences, and is also subject to some degree of inter-subject variability, similar to some subjective image properties. Interestingly, when exposed to the overflow of visual images, subjects have rather consistent tendency to remember or forget the same images \cite{standing1973learning,vogt2007long}. This suggests that subjects encode the same type of visual information, despite great inter-subject variabilities.
% Understanding and predicting memorability may be widely applied in advertisement design, journal cover selection, etc.
% Studying what makes images memorable and how to predict memorability, may have many applications in commercial, advertisement, education and etc.
Recent works \cite{isola2011makes,isola2014makes,khosla2012image,khosla2013modifying,khosla2015understanding} analyze the reason why people have the intuition to remember images, and provide reliable solutions for ranking images by memorability scores. These works are mostly for generic images \cite{isola2011makes,isola2011understanding,khosla2012image,mancas2013memorability,isola2014makes,khosla2015understanding,bylinskii2015intrinsic}, object images \cite{dubey2015makes,han2015learning} and face photographs \cite{bainbridge2012establishing,bainbridge2013intrinsic,khosla2013modifying}. However, it is difficult to dig out the obvious cues relevant to the memorability of natural scene. To date, methods for predicting the visual memorability of natural scene are scarce.

Unlike object-centric images or portraits, subjects cannot clearly clarify which part of natural scene sticks in mind or fades away. To our best knowledge, \cite{Lu2016Predicting} is the only work on memorability of natural scene, which analyzes the relationship between a set of low-level features and their memorability scores. However, the low-level features have limitation in improving the performance of memorability prediction, as verified in this paper. Thus, the approach of \cite{Lu2016Predicting} performs moderately and still has margin with human consistency. In particular, the certain middle- and high-level visual features are not explored in \cite{Lu2016Predicting}, which may help to differentiate memorable and forgettable natural scene images. %For example, Fig. \ref{fig:Figure first} shows the correlation between high-level feature of scene category and memorability of natural scene.
Furthermore, deep neural network (DNN) can enable us to go far in predicting natural scene memorability with high accuracy.
%Unlike object-centric images or portraits, subjects cannot clearly clarify which part of natural scene sticks in mind or fades away. To our best knowledge, \cite{Lu2016Predicting} is the only work on memorability of natural scene, which analyzed the relationship between a set of handcrafted low-level features and their memorability scores. However, the low-level features have limitation in improving the performance of memorability prediction, as verified in this paper. Thus, the approach of \cite{Lu2016Predicting} performs moderately and still has margin with human consistency. In particular, the certain middle- and high-level visual features are not explored in \cite{Lu2016Predicting}, which may help to differentiate memorable and forgettable natural scene images. For example, Figure \ref{fig:Figure first} shows that there exists correlation between high-level feature of scene category and memorability of natural scene. Furthermore, deep neural network (DNN) can enable us to go far in predicting natural scene memorability with high accuracy.

% \begin{figure}[t]
% \begin{center}

% \includegraphics[width=.99\linewidth]{Figure_1}
% \end{center}

% \caption{Image samples from different natural scene categories. The ground truth memorability score is annotated in each image. The number in the bracket represents average score of each category. It can be clearly seen that scene category is correlated with  natural scene memorability.}\label{fig:Figure first}

% \end{figure}

In this paper, we systematically explore what makes natural scene memorable, and propose a DNN based method to predict the memorability of natural scene images. In exploring the connection between image features and natural scene memorability, we make the following contributions, which have potential applications in designing tourism publicity materials, selecting magazine cover, and so forth.

\begin{figure*}[t]
\begin{center}

\includegraphics[width=1\linewidth]{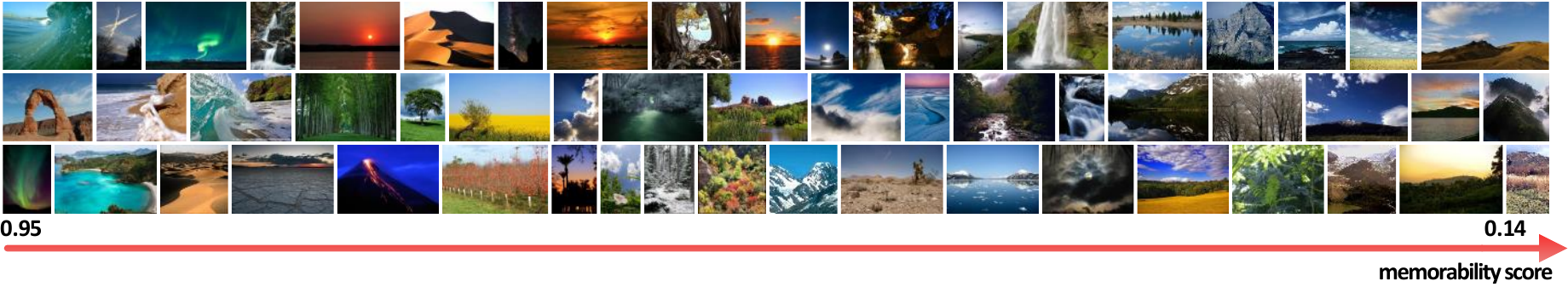}
\end{center}

\caption{Image samples from our LNSIM database. The images above are ranked by their memorability scores, which decrease from left to right.}\label{fig:Figure 1}

\end{figure*}

\textbf{Contributions:} (1) This work stands as the first to introduce a large-scale natural scene image memorability (LNSIM) database, containing 2,632 natural scene images with memorability scores (10 times larger than the previous natural scene memorability database \cite{Lu2016Predicting}).
%The extensive data of LNSIM enable the thorough studies on natural scene memorability and the training of DNN to predict memorability. Moreover, it is surprising that human are highly consistent in remembering natural scenes, similar to human consistency on generic images, object images and face photographs.
(2) We quantify the interplay between natural scene memorability and various visual factors, including deep feature and low-, middle- and high-level (i.e., scene category) features. In particular, we find that scene category is rather correlated with natural scene memorability. %Besides, scene category is verified to have a positive role in promoting the prediction performance of deep feature (i.e., the feature extracted from DNN).
(3) Motivated by the above observations, we propose a DNN based natural scene memorability (DeepNSM) predictor, which outperforms the state-of-the-art methods.

% \vspace{-1em}

\section{Related work}\label{related}

\hspace{1em}\textbf{Memorability of generic images.} Isola \textit{et al.} \cite{isola2011makes} pioneered on the study of image memorability for generic images, and they have shown that memorability is an intrinsic property of an image. They further analyzed how various visual factors influence the memorability of generic images, and utilized global feature combination for memorability prediction \cite{isola2011understanding}. In a more recent study, Khosla \textit{et al.} combined local features with global features to increase the prediction performance. Then Mancas \textit{et al.} \cite{mancas2013memorability} suggested that incorporating the attention-related feature in \cite{isola2011makes} further improves the prediction accuracy.
% the memorability prediction.
Meanwhile,
% Celikkale \textit{et al.} \cite{celikkale2013visual} found that bottom-up and object-level saliency maps capturing information about foreground objects play an important role in memorability prediction. Therefore,
a visual attention-driven
% spatial pooling
approach was proposed in \cite{celikkale2013visual}. Later, Bylinskii \textit{et al.} \cite{bylinskii2015intrinsic} investigated the interplay between intrinsic and extrinsic factors that affect image memorability, and then they developed a more complete and fine-grained model of image memorability.

Recently, DNN has shown splendid achievement in many research areas, e.g., video coding \cite{xu2018reducing} and computer vision \cite{yang2017enhancing,yang2018multi}. Also, several DNN approaches were proposed to estimate image memorability, which significantly improve the prediction accuracy. Specifically, Khosla \textit{et al.} \cite{khosla2015understanding} trained the MemNet on a large-scale database, achieving a splendid prediction performance close to human consistency. In addition, Baveye \textit{et al.} \cite{baveye2016deep} fine-tuned the GoogleNet on the same datatbase of \cite{isola2011makes}, exceeding the performance of handcrafted features mentioned above. They also cast light on the importance of balancing emotional bias, when establishing the memorability-related database.

\textbf{Memorability of faces, objects and natural scene.} %To better understand and predict image memorability,
The study of image memorability on certain targets, like faces, objects and natural scenes, has recently attracted the interests of computer vision researchers \cite{khosla2013modifying,bainbridge2012establishing,bainbridge2013intrinsic,dubey2015makes,Lu2016Predicting}. Bainbridge \textit{et al.} \cite{bainbridge2012establishing} firstly established a database for studying the memorability of human face photographs. They further explored the contribution of certain traits (e.g., kindness, trustworthiness, etc.) to face memorability, but such traits only partly explain facial memorability. Furthermore, \cite{khosla2013modifying} proposed a method to modify the memorability of individual face photographs.

Dubey \textit{et al.} \cite{dubey2015makes}
%were the first to
studied the problem of object memorability and assumed that object categories play an important role in determining object memorability.
%They obtained the memorability scores of all constituent objects possibly appearing in object images
%by subjective experiment.
%Since the splendid performance of DNN is achieved in various recognition tasks,
Then, they utilized the deep feature extracted by conv-net \cite{krizhevsky2012imagenet,jia2014caffe} and ground truth scores of objects to predict object memorability better. Besides, \cite{han2015learning} learned video memorability from brain functional magnetic resonance imaging (fMRI). More recently, Lu \textit{et al.} \cite{Lu2016Predicting} studied the memorability of natural scene on the subset of database in \cite{isola2011makes}. They indicated that the HSV color features perform well on the natural scene memorability, and then they combined the HSV-based feature and other traditional low-level features to predict memorability scores. Nonetheless, only handcrafted low-level features are considered in memorability prediction, which are limited in the prediction accuracy.

\section{Natural scene memorability database}\label{section:2}

As a first step towards understanding and predicting the memorability of nature scene, we build the LNSIM database. Our database with memorability scores is available on line. %(not provided in this manuscript due to the double blind review).

\begin{figure*}[t]
\begin{center}

\includegraphics[width=.99\linewidth]{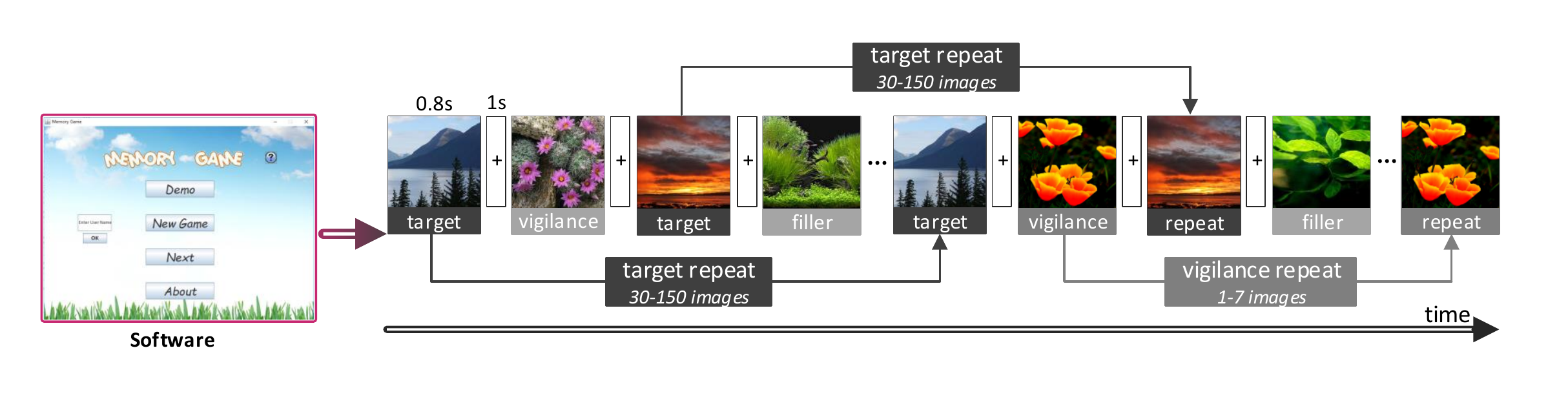}
\end{center}

\caption{The experimental procedure of our memory game. Each level lasts about 5.5 minutes with a total of 186 images. Those 186 images are composed of 66 targets, 30 fillers and 12 vigilance images. The specific time durations for experiment setting are labeled above.}\label{fig:game}

\end{figure*}

% \vspace{-1em}
\subsection{Database establishment}

\textbf{\ \ \ \ Collecting images.} In our LNSIM database, there are in total 2,632 natural scene images. For obtaining these natural scene images, we first selected 6,886 images, which contain natural scenes from the existing databases, including MIR Flickr \cite{Huiskes2008The}, MIT1003 \cite{Judd2009Learning}, NUSEF \cite{NUSEF2010}, SUN \cite{Xiao2010}, affective image database \cite{Machajdik2010Affective}, and AVA database \cite{Murray2012AVA}. Since the natural scene images are hard to be distinguished, 5 volunteers were asked to select the natural scene images from 6,886 images with the following two criteria \cite{Lu2016Predicting}: (1) Each image is with outdoor natural scenes. (2) Each image is only composed of natural scenes, not having any human, animal and man-made object. Afterwards, the images, chosen by at least four volunteers, were included in our LNSIM database. As a result, 2,632 natural scene images were obtained for the LNSIM database, to be scored with memorability. Note that the resolution of these images ranges from 238$\times$168 to 3776$\times$2517. Fig. \ref{fig:Figure 1} shows some example images from our LNSIM database.

\textbf{Subjective experiments.} In our experiment, we adopted the similar way of \cite{khosla2015understanding} to set up a memory game, which was used to quantify the memorability of each image in our LNSIM database. Note that a software is developed for our memory game. In total, 104 subjects (47 females and 57 males) were involved in our memory game. They do not overlap with the volunteers who participated in the image selection. The procedure of our memory game is summarized in Fig. \ref{fig:game}.

In our experiment, there were 2,632 target images, 488 vigilance images and 1,200 filler images, which were unknown to all subjects. Vigilance and filler images were randomly sampled from the rest of 6,886 images. Target images, as stimuli for our experiment, were randomly repeated with a spacing of 35-150 images. Vigilance images were repeated within 7 images, in attempt to ensure that the subjects were paying attention to the game. Filler images were presented for once, such that spacing between the same target or vigilance images can be inserted. After collecting the data, we assigned a memorability score to quantify how memorable each image is, following the way of \cite{khosla2015understanding}. Since the time intervals of repeat on target images were various in our experiment, we regularized the various time intervals to a certain time $T$. In this paper, we set $T$ to be the time duration of displaying 100 images, as the repeat spacing of targets ranges from 35 to 150.

% \textbf{Memorability scores.} On average, we obtained over 80 valid memory results per target image. The average hit rate on target images was 73.7\% with standard deviation (SD) of 14.2\%, running on the experimental results of 104 subjects. Compared with the database of generic images (average score: 67.5\%, SD: 13.6\%), this implies that the subjects indeed concentrated on the game. The average false alarm rate was 8.14\% (SD of 0.81\%). As the false alarm rate was low in comparison with the hit rate, it eliminates the possibility of hitting correct images only by chance. Thus, our data can reliably reflect the memorability of natural scene images.

% After collecting the data, we assigned a memorability score to quantify how memorable each image is, following the way of \cite{khosla2015understanding}. Since the time intervals of repeat on target images were various in our experiment, we regularized the various time intervals to a certain time $T$. In this paper, we set $T$ to be the time duration of displaying 100 images, as the repeat spacing of targets ranges from 35 to 150.

\textbf{Training and test sets.} In this paper, we refer to the scores collected by the aforementioned memory game as the ``ground truth'' memorability for each image. The 2,632 natural scene images with their ground truth memorability scores are randomly divided into the non-overlapping training and test sets. The training set contains 2,200 images, and the remaining 432 images are used for test.
% , which validates the availability of measurements on two groups
\begin{figure}[t]
\begin{center}
\includegraphics[width=1.0\linewidth]{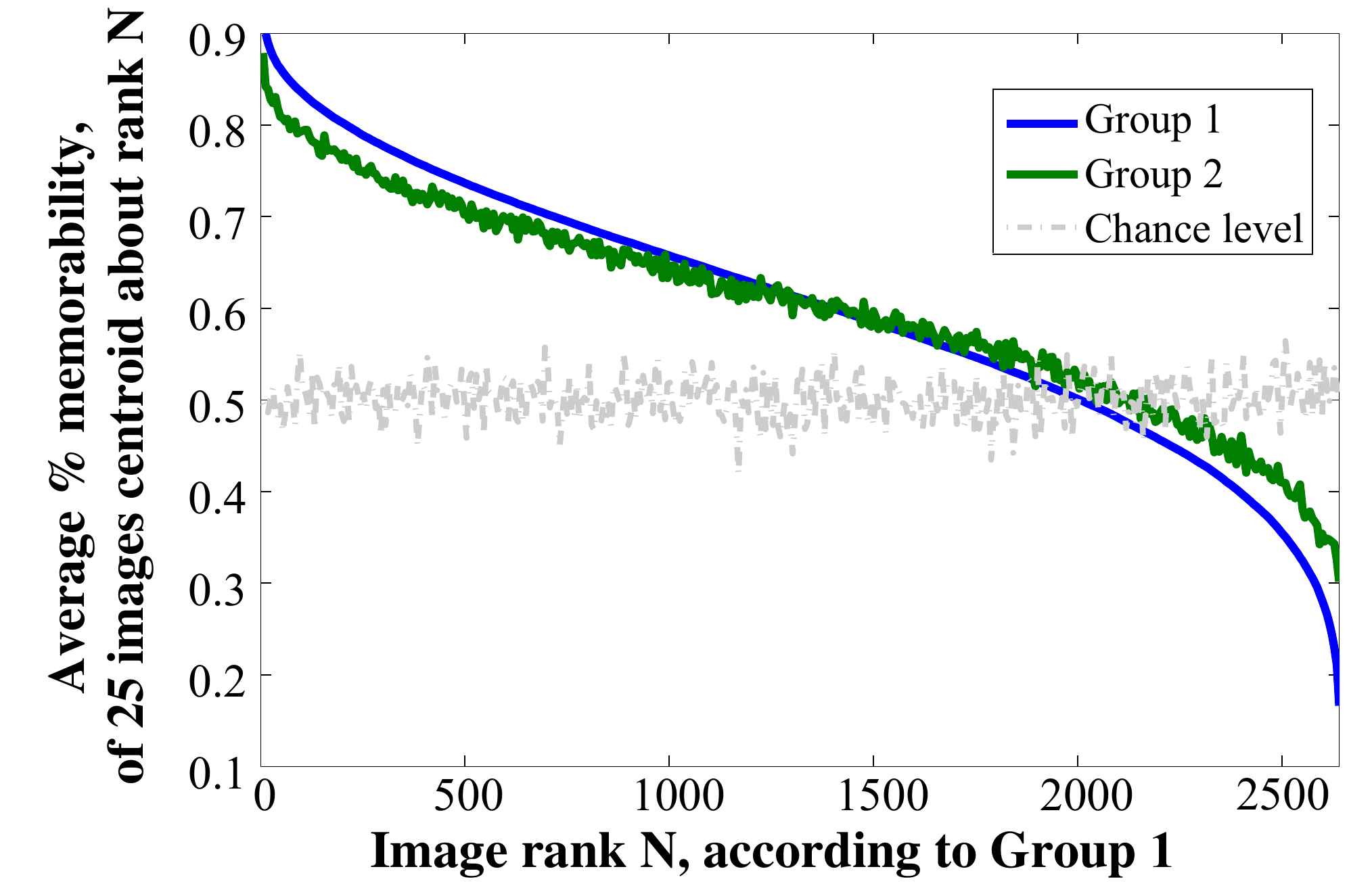}
\end{center}

\caption{Measure of human consistency in natural scene memorability. The chance line is provided by allocating random prediction scores as a reference.
% The memorability scores are derived from two groups of subjects. Images are ranked by memorability scores of subjects in Group 1, and then the curves plot the average memorability scores of Group 1 vs. Group 2. For clarity, we convolute the resulting plots with a length-6 box filter along with the horizontal axis.
% The chance line is provided by allocating random prediction scores as a reference.
}\label{fig:img3}

\end{figure}

\subsection{Human consistency}

To evaluate human consistency, we randomly split subjects into two independent halves (i.e., Groups 1 and 2).
%, and then measured the correlation between memorability scores of these two groups.
%We examined consistency with a variant of correlation measurement:
%We sorted all the 2,632 images by their scores of the first half of subjects, and calculated the corresponding cumulative average memorability scores, according to the second half of subjects. In this fashion,
Fig. \ref{fig:img3} plots the memorability scores of these two groups, %and averaged over 25 random splits,
in which the scores of Group 1 are set as benchmark. Note that the horizontal axis ranks the images with the memorability scores of Group 1 in the decreasing order. As shown in Fig. \ref{fig:img3}, there exists high consistency between two groups of subjects, especially compared to that of the random prediction. We further quantified the human-to-human consistency by measuring the Spearman rank correlation coefficient (SRCC, denoted by $\rho$). The SRCC on the LSNIM database is 0.78 between two sets of scores, indicating that humans are highly consistent in remembering natural scene images.
%measured over 25 random splits. Compared with \cite{isola2014makes}, SRCC measured on natural scene image set is a little higher than that calculated in generic image set ($\rho=0.75$). Furthermore, we selected the top 100 most memorable images with average score of 83.5\% marked by Group 1, and then we obtained their another average scores of 79.4\% from second half of subjects (denoted as Group 2). The above results indicate that the individual differences add noise to estimation; nonetheless, different subjects tend to remember or forget the same images.
%To conclude, humans are highly consistent in remembering natural scene images.
This validates our memory game in obtaining ground truth memorability scores. This also indicates that the memorability of natural scene can potentially be predicted with high accuracy. In the next section, we study the various factors that make natural scene memorable.

\section{Analysis on natural scene memorability}\label{analysis}

In this section, we mine our LNSIM database to better understand how natural scene memorability is influenced by the low-, middle- and high-level handcrafted features and the learned deep feature.

\subsection{Low-level features vs. memorability}

On the basis of predecessors \cite{isola2011makes,isola2014makes,khosla2015understanding}, it has been verified that low-level features, like pixels, SIFT \cite{lazebnik2006beyond} and HOG2$\times$2 \cite{felzenszwalb2010object}, have impact on memorability of generic images. Here, we investigate whether these low-level features still work on natural scene image set as well. To this end, we train a support vector regression (SVR) for each low-level feature using our training set to predict memorability, and then evaluate the SRCC of these low-level features with memorability on the test set. %The histogram intersection kernels\footnote{Note that we traverse all possible kernels for each feature, and select the one with the best performance.} are utilized for these features. Note that
% the train set is used to tune the hyper-parameters for the kernels and
%these low-level features are extracted in the same manner as \cite{khosla2012memorability}.
Table \ref{tab:globalfeats} reports the results of SRCC on natural scenes, with SRCC on generic images \cite{isola2014makes} as the baseline. It is evident that pixels ($\rho=0.08$), SIFT ($\rho=0.28$) and HOG2$\times$2 ($\rho=0.29$) are not as effective as expected on natural scene, especially compared to generic images.For example, the feature of SIFT has capacity to reflect the memorability of generic images to a certain degree with $\rho=0.41$, but its SRCC decreases to 0.28 on natural scene images. This suggests that the low-level features cannot effectively characterize the visual information for remembering natural scenes. Then, we additionally train an SVR on a kernel sum of these low-level features, achieving a rank correlation of $\rho=0.33$. This is less than the SRCC ($\rho=0.45$) of feature combination for generic images.

\begin{table}[t]
\footnotesize
  \centering

    \caption{The correlation $\rho$ between low-level features and natural scene memorability.}\label{tab:globalfeats}
    \vspace{-1em}
    \begin{tabular}{cccccc}
    \toprule
  Database       &pixels&SIFT \cite{lazebnik2006beyond}&HOG\cite{felzenszwalb2010object}&Combination&Human\\
    \midrule
    Our LNSIM    & 0.08    & 0.28 & 0.29 & 0.33 & 0.78 \\
    \midrule
    Generic images \cite{isola2014makes}     & 0.22    & 0.41 & 0.43 & 0.45 & 0.75 \\
    \bottomrule
    \end{tabular}

\end{table}

\subsection{Middle-level features vs. memorability}

The middle-level feature of GIST \cite{oliva2001modeling} describes the spatial structure of an image. Previous works \cite{isola2011makes,isola2014makes} mentioned that GIST is correlated with memorability on generic images.
% ($\rho=0.38$, see Table \ref{tab:midfeats}). In view of this observation, we train an SVR predictor with a RBF kernel for quantifying the correlation between the GIST feature and memorability of natural scene. Note that the training set is used to tune the hyper-parameters for the kernels.
However, Table \ref{tab:midfeats} shows that the SRCC of GIST is only 0.23 for natural scene, much less than $\rho=0.38$ of generic images. This illustrates that structural information provided by the GIST feature is less effective for predicting memorability scores on natural scenes.

% \begin{figure}[t]
% \begin{center}
% % \vspace{-。5em}
% \includegraphics[width=0.60\linewidth]{saliency}
% \end{center}
% \vspace{-1em}
% \caption{Averaged saliency map on images of high, medium and low memorability.}\label{fig:sa}
% \vspace{-1.5em}
% \end{figure}

Intuitively, the region that attracts visual attention in a natural scene may affect image memorability. The work of \cite{mancas2013memorability,celikkale2013visual} attempted to explain memorability of generic images using visual attention-driven features. To quantify the correlation of visual attention with memorability on natural scenes, we apply three state-of-the-art models of visual attention (i.e., PQFT \cite{guo2010novel}, SalGAN \cite{pan2017salgan}, DVA \cite{wang2018deep}) to extract saliency maps. All saliency maps are scaled to $256\times256$, and then we densely sample these maps of each model in a regular grid, resulting in a feature of dimension 1024. Similar to other features, we utilize an SVR predictor to measure the SRCC of the saliency features. Note that the RBF kernel is chosen for the saliency features.% We further split our LNSIM database into three classes: high memorability (score $\geq0.7$), medium memorability ($0.7>$ score $\geq0.4$) and low memorability (score $<0.4$). Figure \ref{fig:sa} demonstrates the averaged saliency maps of each class.

% Additionally, Table \ref{tab:midfeats} compares the correlation of saliency features with memorability of natural scene images and generic images. The results indicate that saliency features extracted from PQFT are more effective in natural scenes than the other models. Such conclusion can also be found from Figure \ref{fig:sa}. Conversely, as shown in Table \ref{tab:midfeats}, the saliency features extracted from PQFT are worst for predicting memorability of generic images. This suggests that when predicting natural scene memorability, frequency domain saliency model (PQFT) performs better than other pixel domain models.

% In general, middle-level features have modest connection with memorability of natural scenes. In some degree, the prediction performance of middle level features inferiors to some high-dimension low-level features.

\subsection{High-level feature vs. memorability}

\begin{figure*}[t]
\begin{center}
\subfigure{\includegraphics[width=.95\linewidth]{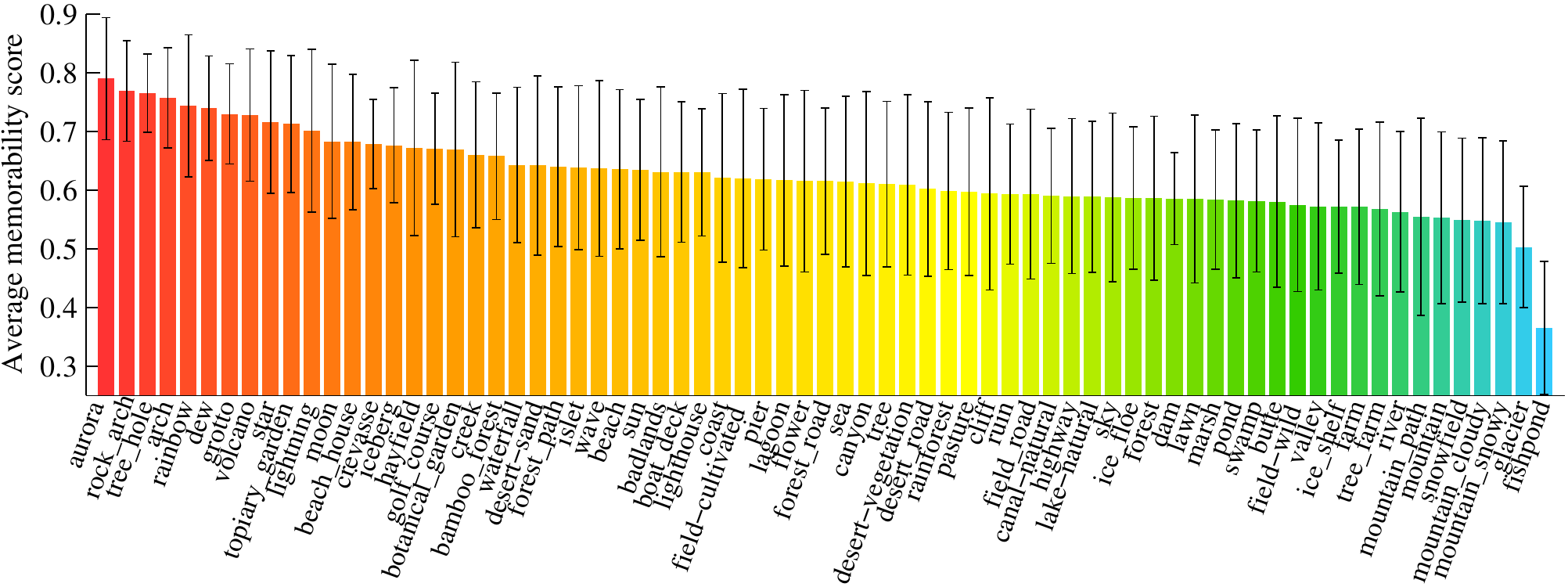}}
% \subfigure[]{\includegraphics[width=0.16\linewidth]{untitled}}
\end{center}

\caption{Comparison of average memorability score and standard deviation of each scene category.}\label{fig:scenecat}

\end{figure*}

There is no salient object, animal or person in natural scene images, such that scene semantics, as a high-level feature, may be interpreted as something relevant to landform, celestial body, botany and so on. Similar to object detection, we use scene category attribute to characterize scene semantics of each natural scene image. According to WordNet taxonomy \cite{miller1995wordnet}, our LNSIM database includes 71 scene categories (badlands, coast, desert, etc.), which are non-overlapped with each other. To obtain the ground truth of scene category, we design two experiments to annotate scene category for 2,632 images in our database.

\begin{itemize}
\item Task 1 (Classification Judgment): We asked 5 participants to indicate which scene categories an image has. A random image query was generated for each participant. We showed an image and all scene categories at a time. Participants had to choose proper scene category labels to interpret scene stuff for each image.
\item Task 2 (Verification Judgment): We further ran a separate task on the same set of images by recruiting another 5 participants after Task 1. For a given category name, a single image was shown centered in the screen, with a question like ``\textit{is this a coast scene?}'' The participants were asked to provide a binary answer to the question for each image. The default answer was set to ``No'', and the participants can check the box of image index to set ``No'' to ``Yes''.
\end{itemize}

We annotated all images with categories through the majority voting over Task 1 and Task 2. Specifically, Task 1 completed the natural scene category annotation initially, while Task 2 amended the results of Task 1. For each image of our database, we determined its scene categories according to the results of Task 2. In this way, the scene categories of all 2,632 images in the LNSIM database were obtained, taking account of 10 participants' selection. Note that all 10 participants did not attend the memory game, and one image may have more than one category in our database. Additionally, the rate of choosing ``Yes'' in Task 2 is $81\%$ among the 5 participants. This indicates that different annotators are consistent in classifying scene segmentation.

% Additionally, we calculated the average classification correlation coefficient of Task 2 (0.81) on 71 scene categories against the baseline results of Task 1.

Afterwards, we test the memorability prediction performance of scene category on the LNSIM database. An SVR predictor with the histogram intersection kernel is trained for scene category. The scene category attribute achieves a good performance of SRCC ($\rho=0.38$), outperforming the results of low-level feature combination. This suggests that high-level scene category is an obvious cue of quantifying the natural scene memorability. We further analyze the connection between different scene categories and natural scene memorability. To this end, we use the mean and SD values of memorability scores in each category to quantify such relationship. As shown in Figure \ref{fig:scenecat}, the horizontal axis represents scene categories in the descending order of corresponding average memorability scores. The average score ranges from 0.79 to 0.36, giving a sense of how memorability changes across different scene categories. The distribution in Figure \ref{fig:scenecat} indicates that some unusual classes like aurora tend to be more memorable, while usual classes like mountain are more likely to be forgotten. This is possibly due to the frequency of each category appears in daily life.

% \begin{table*}[t]
% \footnotesize
%   \centering
% %   \vspace{-.5em}
%     \caption{The correlation $\rho$ between combined features and natural scene memorability.}\label{tab:deep}
%     \vspace{-1em}
%     \begin{tabular}{ccccccccccc}
%     \toprule
%     & - &pixels&SIFT \cite{lazebnik2006beyond}&HOG\cite{felzenszwalb2010object} &GIST \cite{oliva2001modeling}&HSV-based \cite{Lu2016Predicting}&PQFT \cite{guo2010novel}&SalGAN \cite{pan2017salgan}&DVA \cite{wang2018deep}&Scene category\\
%     \midrule
%     Deep feature   & 0.44  & 0.44  & 0.44 & 0.44 & 0.44 & 0.44 & 0.44 & 0.44 & 0.44 & \textbf{0.46} \\
% %     \midrule
% %     Generic images database \cite{isola2014makes}     & 0.38    & 0.15  & 0.27  &  0.30 \\
%     \bottomrule
%     \end{tabular}
%     \vspace{-.8em}
% \end{table*}

\begin{table}[t]
\footnotesize
  \centering

    \caption{The correlation $\rho$ between middle-level features and natural scene memorability.}\label{tab:midfeats}

    \begin{tabular}{ccccc}
    \toprule
  Database       &GIST \cite{oliva2001modeling}&PQFT \cite{guo2010novel}&SalGAN \cite{pan2017salgan}&DVA \cite{wang2018deep}\\
    \midrule
    Our LNSIM     & 0.23  & 0.25  & 0.20 & 0.20 \\
    \midrule
    Generic images \cite{isola2014makes}     & 0.38    & 0.15  & 0.27  &  0.30 \\
    \bottomrule
    \end{tabular}

\end{table}

\subsection{Deep feature vs. memorability}\label{deep}
DNN models have recently been shown to achieve splendid results in various tasks in the field of computer vision, among which \cite{khosla2015understanding,baveye2016deep,dubey2015makes} utilized DNN to predict image memorability. To dig out how deep feature influences the memorability of natural scene, we fine-tuned MemNet\footnote{MemNet is proposed for predicting the memorability scores of generic images.} \cite{khosla2015understanding} on our training set of LNSIM database, using the Euclidean distance between the predicted and ground truth memorability scores as the loss function. We extract the output of the last hidden layer as the deep feature (dimension: 4096).
% Note that the deep feature consists of hierarchical learned features from low- to high-level. The histogram intersection kernel is employed in an SVR predictor for the deep feature.
To evaluate the correlation between the deep feature and natural scene memorability, similar to above handcrafted features, an SVR predictor with histogram intersection kernel is trained for the deep feature. The SRCC of deep feature is 0.44, exceeding all handcrafted features. It is acceptable that DNN indeed works well on predicting the memorability of natural scene, as deep feature shows a rather high prediction accuracy. Nonetheless, there is no doubt that the fine-tuned MemNet also has its limitation, since it still has gap to human consistency ($\rho=0.78$).

% We further combine the deep feature with each of the aforementioned low-, middle- and high-level feature, to explore whether such combination is able to improve the prediction accuracy. The SRCC value stays almost unchanged, when combining each low/middle-level feature with deep feature. It is probably because DNN has the ability to extract hierarchical features of different levels, leading to the ineffectiveness of combining low- and middle-level features. However, the scene category, as a high-level feature, helps to increase the SRCC of the deep feature from $\rho=0.44$ to $\rho=0.46$. This may be due to the fact that the architecture of MemNet is too simple to adequately learn the high-level feature, so that combining with the high-level feature is advantageous for the deep feature to predict natural scene memorability. Motivated by this, we propose a scene category based DNN approach to predict the memorability natural scene images in the next section.

% The results show the mutual effect between deep feature and other handcrafted feature is rather small. It is noticeable that the high level feature of scene category play a slightly positive role in promoting the prediction performance of deep feature. This suggests that scene category can serve as supplement for deep feature, which contribute to predicting the memorability of natural scene. Motivated by this, we propose a DNN based approach to predict the memorability scores of natural scene images, and more details are discussed in the following.

\section{Predicting natural scene memorability}

In above, we have analyzed how individual and combined visual features reflect natural scene memorability.
% Such analysis can serve as a stepping stone in the direction of designing a method to automatically predict the memorability of natural scene images.
Accordingly, the deep feature and scene category is most effective in predicting memorability of natural scene. Therefore, we propose an end-to-end DeepNSM method, which exploits both deep and category-related features for predicting the natural scene memorability.
\begin{figure*}[t]
\begin{center}

\includegraphics[width=0.95\linewidth]{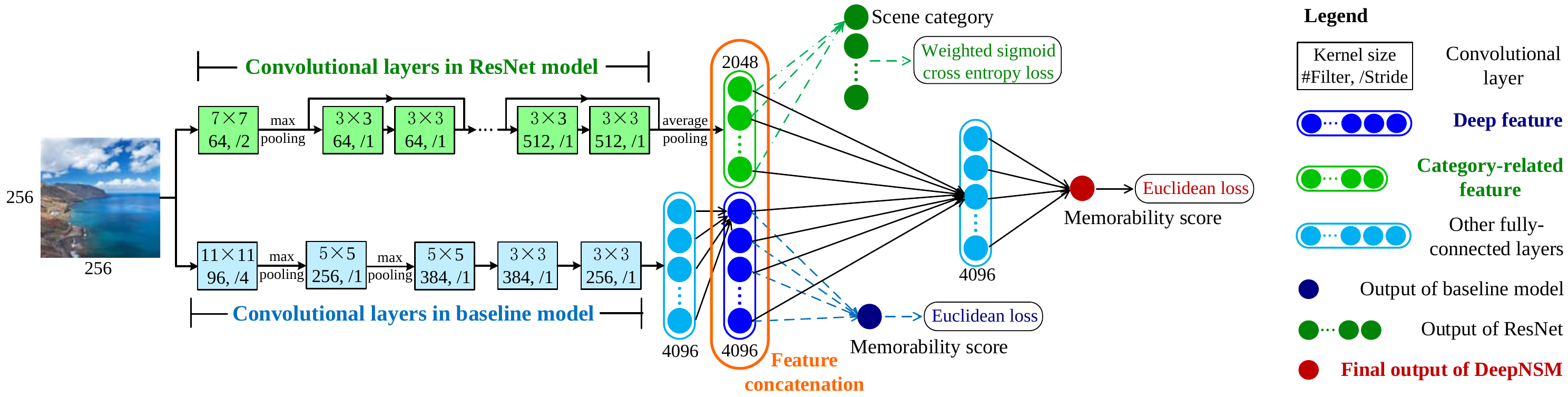}
\end{center}
\caption{Architecture of our DeepNSM model.}\label{fig:net}
\end{figure*}

\subsection{DeepNSM: DNN for natural scene memorability}

As discussed in Section \ref{analysis}, MemNet, which is fine-tuned on our training set, outperforms all the low-, middle- and high-level visual features. Hence, the fine-tuned MemNet model serves as the baseline model on predicting natural scene memorability. Besides, Section \ref{analysis} shows that the high-level feature of scene category is rather correlated to natural scene memorability. By contrast, the low- and middle-level visual features are with poor performance. Therefore, in the proposed DeepNSM architecture, the deep feature
%\footnote{Although the category-related feature is also obtained from DNN, to keep consistent with the previous context, ``deep feature'' particularly refers to the 4096 dimensional feature extracted from the baseline model.}
is concatenated with category-related feature to accurately predict the memorability of natural scene images. Note that the ``deep feature'' refers to the 4096-dimension feature extracted from the baseline model.

% In order to extract scene category features, we fine-tune the ResNet \cite{he2016deep} on our training set according to the ground-truth labels of scene categories.
% The architecture of the ResNet is shown in Figure \ref{fig:net}.

\textbf{Extracting category-related feature.} In DeepNSM, ResNet \cite{he2016deep} is applied to extract the category-related feature. We first initialize ResNet with the pre-trained model on ImageNet \cite{Deng2009ImageNet}. Then, 33,000 natural scene images selected from the database of Places \cite{zhou2017places} are adopted to fine-tune the ResNet model. Finally, it is further fine-tuned on our training set according to the ground truth labels of scene category. Note that different from the databases of ImageNet and Places, whose labels are one-hot, each image in our LNSIM database may contain several categories. As such, it is a multi-label classification task. Thus, the weighted sigmoid cross entropy is utilized as the loss function, instead of softmax loss in \cite{he2016deep}. The fine-tuned ResNet can be seen as a category-related feature extractor. The output of the hidden fully-connected layer in ResNet is used as the extracted category-related feature. See Figure \ref{fig:net} for more details.

\textbf{The proposed architecture.} Finally, the architecture of our DeepNSM model is presented in Figure \ref{fig:net}. In our DeepNSM model, the aforementioned category-related feature is concatenated with the deep feature obtained from the baseline model. Based on such concatenated feature, additional fully-connected layers (including one hidden layer with dimension of 4096) are designed to predict the memorability scores of natural scene images. In training, the layers of the baseline and ResNet models are initialized by the individually pre-trained models, and the added fully-connected layers are randomly initialized. The whole network is jointly trained in an end-to-end manner, using the Adam \cite{Kingma2014Adam} optimizer with the Euclidean distance adopted as the loss function.

Note that although some existing memorability prediction works \cite{isola2014makes,isola2011makes} also take image category into consideration, they only apply the manually classified ground truth category information. To the best of our knowledge, our work is the first attempt to automatically extract the category-related feature by DNN in predicting memorability. The advantage is two fold: (1) The image memorability can be predicted without any manual annotation; (2) It is able to achieve the end-to-end training of the DNN model.

\subsection{Performance evaluation}

Now, we evaluate the performance of our DeepNSM model on predicting natural scene memorability in terms of SRCC ($\rho$). Our DeepNSM model is tested on both the test set of our LNSIM database and the NSIM database introduced in \cite{Lu2016Predicting}. The SRCC performance of our DeepNSM model is compared with the state-of-the-art memorability prediction methods, including MemNet \cite{khosla2015understanding}, MemoNet \cite{baveye2016deep} and Lu \textit{et al.} \cite{Lu2016Predicting}. Among them, MemNet \cite{khosla2015understanding} and MemoNet \cite{baveye2016deep} are the latest DNN methods for generic images, which beat the conventional methods using handcrafted features. Lu \textit{et al.} \cite{Lu2016Predicting} is a state-of-the-art method for predicting natural scene memorability.
% Note that we do not compare with Isola \textit{et al.} \cite{isola2014makes}, since it relies on several manual annotated features which is not available in our LNSIM database.
% More importantly, it is proved that the state-of-the-art DNN methods outperform the method of \cite{isola2014makes}, which uses traditional SVR.

\begin{table}[!t]
\footnotesize
  \centering
    \caption{The SRCC ($\rho$) performance of our DeepNSM and compared methods.}\label{tab:results}
    \begin{tabular}{ccccc}
    \toprule
      Database&MemNet \cite{khosla2015understanding}&MemoNet \cite{baveye2016deep} &Lu \textit{et al.} \cite{Lu2016Predicting}&Our DeepNSM \\
    \midrule
    Our LNSIM     & 0.43  & 0.39  & 0.19 & \textbf{0.58} \\
    \midrule
    NSIM \cite{Lu2016Predicting}      & 0.40   & -*& 0.47  & \textbf{0.55} \\
    \bottomrule
    \\
    \multicolumn{5}{p{8cm}}{* MemoNet is not tested on the NSIM database, since the NSIM database is completely included in the training set of MemoNet.}\\
    \end{tabular}
\end{table}

\textbf{Comparison with latest DNN methods.} Table \ref{tab:results} shows the SRCC performance of our DeepNSM and the three compared methods. Our DeepNSM successfully achieves the outstanding SRCC performance, i.e., $\rho=0.58$ and $0.55$, over the LNSIM and NSIM \cite{Lu2016Predicting} databases, respectively. It significantly outperforms the state-of-the-art DNN methods, MemNet \cite{khosla2015understanding} and MemoNet \cite{baveye2016deep}.
%MemNet only has $\rho=0.43$ on our LNSIM test set and $\rho=0.40$ on the NSIM database. The SRCC of MemoNet is $\rho=0.39$ on our LNSIM test set. Note that for fair comparison,
Note that we do not test MemoNet on NSIM database, because it is completely included in training the MemoNet model. The above results demonstrate the effectiveness of our DeepNSM in predicting natural scene memorability. It is worth pointing out that as claimed in \cite{khosla2015understanding} and \cite{baveye2016deep}, both MemNet and MemoNet methods are able to reach $\rho=0.64$ on generic images. Nevertheless, their performance severely degrades on natural scenes, and thus validates the difference of factors influencing the memorability of generic and natural scene images. Besides, it also reflects the difficulty to accurately predict natural scene memorability. In summary, our DeepNSM outperforms the state-of-the-art DNN methods on predicting natural scene memorability, and more importantly, makes up the shortcomings of these generic image methods.

\textbf{Comparison with the latest natural scene method.} We compare our DeepNSM model with the latest method \cite{Lu2016Predicting}, which is designed for predicting natural scene memorability. As shown in Table \ref{tab:results}, our DeepNSM model outperforms the method of \cite{Lu2016Predicting} on both LNSIM and NSIM databases. Moreover, compared with the database of NSIM \cite{Lu2016Predicting} ($\rho=0.47$), the SRCC of \cite{Lu2016Predicting} obviously reduces on our LNSIM database ($\rho=0.19$). On the contrary, our DeepNSM model achieves comparable performance on both databases. This shows the good generalization capacity of our DeepNSM model, which benefits from the large scale training set of our LNSIM database.

\section{Conclusion}
In this paper, we have investigated the memorability of natural scene from data-driven perspective. Specifically, we established the LNSIM database for analyzing human memorability on natural scene. In exploring the correlation of memorability with low-, middle- and high-level features, we found that high-level feature of scene category plays an important role in predicting the memorability of natural scene. %In addition, deep feature shows a positive impact on promoting the prediction performance of natural scene.
Accordingly, we proposed the DeepNSM method for end-to-end predicting natural scene memorability, much better than other state-of-the-art approaches. The analysis of feature in compression domain, such as rate distortion \cite{li2017optimal}, is an interesting future work.
%In this paper, we have investigated the memorability of natural scene from data-driven perspective. Specifically, we established the LNSIM database that helps to study and analyze the human memorability on natural scene in depth. In exploring the correlation of memorability with low-, middle- and high-level features, we found that high-level feature of scene category plays an important role in predicting the memorability of natural scene. In addition, deep feature shows a positive impact on promoting the prediction performance of natural scene. Accordingly, we proposed the DeepNSM method for end-to-end predicting natural scene memorability. The experimental results showed that our DeepNSM model advances the state-of-the-art in memorability prediction of natural scene images.
% Moreover, we studied the strengths and weakness of the DeepNSM model. To break through the weakness of our DeepNSM model, other features affecting memorability of natural scene need to be found and incorporated in our model, as an interesting future work.
%Moreover, the case study is conducted to dig out the reasons of accurate and inaccurate prediction, which shows the promising research direction in the future.

\section*{ACKNOWLEDGMENTS}

This work was supported by the National Nature Science Foundation
of China under Grant 61573037 and by the Fok Ying Tung
Education Foundation under Grant 151061.